\documentclass{article}
\usepackage{amsfonts}
\usepackage{algorithmic}
\usepackage{verbatim}
\usepackage{mathtools}
\usepackage{graphicx}
\usepackage{tikz}

\usepackage{stmaryrd}
\usepackage{booktabs}
\usepackage{amssymb}
\usepackage{amsmath}
\usetikzlibrary{positioning}
\usepackage{caption}
\usepackage{subcaption}
\usepackage{graphicx}
\usepackage{apacite}

\usepackage[a4paper, total={6in, 8in}]{geometry}

\DeclareMathOperator*{\argmax}{arg\,max}

\usetikzlibrary{arrows.meta}
\tikzset{%
	>={Latex[width=2mm,length=2mm]},
	base/.style = {rectangle, rounded corners, draw=black,
		minimum width=4cm, minimum height=1cm,
		text centered, font=\sffamily},
	activityStarts/.style = {base, fill=blue!30},
	startstop/.style = {base, fill=red!30},
	activityRuns/.style = {base, fill=green!30},
	process/.style = {base, minimum width=2.5cm, fill=orange!15,
		font=\ttfamily},
}

\makeatletter
\newcommand{\xmapsfrom}[2][]{%
	\ext@arrow3095\leftarrowfill@{#1}{#2}\mapsfromchar
}
\makeatother

\newtheorem{algo}{Algorithm}{\bf}{\it}

\begin{document}

\title{Smart Containers With Bidding Capacity: A Policy Gradient Algorithm for Semi-Cooperative Learning}

\author{W.J.A. van Heeswijk}
\maketitle              
\begin{abstract}
Smart modular freight containers -- as propagated in the Physical Internet paradigm -- are equipped with sensors, data storage capability and intelligence that enable them to route themselves from origin to destination without manual intervention or central governance. In this self-organizing setting, containers can autonomously place bids on transport services in a spot market setting. However, for individual containers it may be difficult to learn good bidding policies due to limited observations. By sharing information and costs between one another, smart containers can jointly learn bidding policies, even though simultaneously competing for the same transport capacity. We replicate this behavior by learning stochastic bidding policies in a semi-cooperative multi agent setting. To this end, we develop a reinforcement learning algorithm based on the policy gradient framework. Numerical experiments show that sharing solely bids and acceptance decisions leads to stable bidding policies. Additional system information only marginally improves performance; individual job properties suffice to place appropriate bids. Furthermore, we find that carriers may have incentives not to share information with the smart containers. The experiments give rise to several directions for follow-up research, in particular the interaction between smart containers and transport services in self-organizing logistics.
\end{abstract}
\section{\label{sec:introduction1}Introduction}
The logistics domain is increasingly moving towards self-organization, meaning that freight transport is planned without direct human intervention. The Physical Internet is often considered as the ultimate form of self-organizing logistics, having smart modular containers equipped with sensors and intelligence able to interact with their surroundings and to route themselves \cite{vanheeswijk2019b}. Due to the standardized shapes of the containers, they can easily be combined into full truckloads and be decomposed with equal ease. The concept also suggests that the system should be able to function without a high degree of central governance, rather converging to an organically functioning system by itself. Moreover, it is more efficient than traditional logistics systems, being able to dynamically respond to disruptions and opportunities in the logistics system utilizing intelligent decision-making policies. It is this notion of autonomy and self-organizing systems that inspired the present study.

We model smart containers as independent job agents that -- on behalf of their shippers -- are able to place a bid on the transport service that they wish to use. In a dynamic setting, the bid price should depends on the state of the system. Rather than having the fixed contract prices that preside in contemporary transport markets, dynamic bidding mimics financial spot markets that constantly balance demand and supply. For instance, if a warehouse holds relatively few containers waiting for transport, a low bid may suffice to get accepted for the transport service, whereas higher bids might be required during busy times. Additionally, there is also an anticipatory element involved in the bidding decision. Assuming each container has a given due date, the bidding strategy should also take into account the probability of future bids getting accepted. 

The optimal bidding strategy may be influenced by many factors. We want to have a policy that provides us with the bid that minimizes expected transport costs given the current state of the system. Such an optimal policy is very difficult to derive analytically, but may be approximated by means of reinforcement learning. However, as each job terminates upon delivery of the smart container, lifespans of individual jobs are very limited. As the quality of a learned bidding policy to a large extent depends upon the number of observations made, it is therefore difficult to learn policies individually. Semi-cooperative learning \cite{boukhtouta2011} might alleviate this problem; even though all endeavoring to minimize their own costs rather than striving towards a common goal, smart containers can share observations to jointly learn better bidding policies that benefit the individual agents. On the other hand, if competing containers are aware of the exact bidding strategies of other containers, they may easily be countered. A fully deterministic policy might therefore not be realistic, we explore whether a stochastic policy yields sensible bidding decisions. Another question that we seek to answer is whether sharing additional information (other than bid prices and acceptance) helps in improving bidding policies. System information (e.g., total container volume in warehouse) may allow for better bids, but the competing containers also utilize this information for the same purpose.

The contribution of this paper is as follows. First, we explore a setting in which smart containers may place bids on transport capacity; to the best of our knowledge this topic has not been studied before from an operations research perspective. In particular, we aim to provide insights into the drivers that determine the bid price and the effects of information sharing on policy quality. Second, we present a policy gradient reinforcement learning algorithm to learn stochastic bidding policies, aiming to mimic a reality in which competing smart containers may deviate from jointly learned policies. Due to the explorative nature of the paper, we present a simplified problem settings involving a single transport service with fixed capacity that operates on the real line. The focus is on the basic mechanisms that govern bidding dynamics absent regulation and centralized control.

\section{Literature}
This literature overview is structured as follows. First, we discuss the role of smart containers in the Physical Internet. Second, we highlight several studies on reinforcement learning in the Delivery Dispatching Problem, which links to our problem from a carrier's perspective. Third, we discuss studies that address the topic of bidding in freight transport.

The inspiration from this paper stems from the Physical Internet paradigm.  We refer to the seminal works of Montreuil \cite{montreuil2011,montreuil2013} for a conceptual outline of the Physical Internet, thoroughly addressing the foundations of the Physical Internet. It envisions an open market at which logistics services are offered, stating that (potentially automated) interactions between smart containers and other constituents of the Physical Internet determine routing decisions. Sallez \textit{et al.} \cite{sallez2016} stress the active role that smart containers have, being able to communicate, memorize, negotiate, and learn both individually and collectively. Ambra \textit{et al.} \cite{ambra2019} present a recent literature review of work performed in the domain of the Physical Internet. Interestingly, their overview does not mention any paper that defines the smart container itself as the targeted actor. Instead, existing works seem to focus on traditional actors such as carriers, shippers and logistics service providers, even though smart containers supposedly route themselves in the Physical Internet.

The problem studied in this paper is related to the Delivery Dispatching Problem \cite{minkoff1993}, which entails dispatching decisions from a carrier's perspective. In this problem setting, transport jobs arrive at a hub according to some external stochastic process. The carrier subsequently decides which subset of jobs to take, also considering future jobs that arrive according to the stochastic process. The most basic instances may be solved with queuing models, but more complicated variants quickly become computationally intractable, such that researchers often resort to reinforcement learning to learn high-quality policies. We highlight some recent works in this domain. Klapp \textit{et al.} \cite{klapp2018} develop an algorithm that solves the dispatching problem for a transport service operating on the real line. Van Heeswijk \& La Poutr{\'e} \cite{vanheeswijk2018b} compare centralized and decentralized transport for networks with fixed line transport services, concluding that decentralized planning yields considerable computational benefits. Van Heeswijk \textit{et al.} \cite{vanheeswijk2015,vanheeswijk2019} study a variant that includes a routing component, using value function approximation to find policies. Voccia \textit{et al.} \cite{voccia2019} solve a variant that includes both pickups and deliveries.

We highlight some works on optimal bidding in freight transport; most of these studies seem to adopt a viewpoint in which competing carriers bid on transport jobs. For instance, Yan \textit{et al.} \cite{yan2018} propose a particle swarm optimization algorithm used by carriers to place bids on jobs. Miller \& Nie \cite{miller2019} present a solution that emphasizes the importance of integration between carrier competition, routing and bidding. Wang \textit{et al.} \cite{wang2019} design a reinforcement learning algorithm based on knowledge gradients to solve for a bidding structure with a broker intermediating between carriers and shippers. The broker aims to propose a price that satisfies both carrier and shipper, taking a percentage of accepted bids as its reward. In a Physical Internet context, Qiao \textit{et al.} \cite{qiao2019} model hubs as spot freight markets where carriers can place bids on transport bids. To this end, they propose a dynamic pricing model based on an auction mechanism.

\section{Problem definition}
This section formally defines our problem in the form of a Markov Decision Process (MDP) model. The model is designed from the perspective of a modular container -- denoted as a job $\boldsymbol{j}$ -- that aims to minimize its expected shipping costs over a finite discretized time horizon $\mathcal{T}_{\boldsymbol{j}}$, with each decision epoch $t \in \mathcal{T}_{\boldsymbol{j}}$ representing a day on which a bid for a capacitated transport service (the carrier) may be placed. In addition to this job-dependent time horizon, we define a system horizon $\{0,\ldots,T\}$ with corresponding decision epochs denoted by $t^\prime$. Thus, we use $t$ when referring to the individual job level and $t^\prime$ for the system level.

\begin{figure}
	\includegraphics[width=\textwidth]{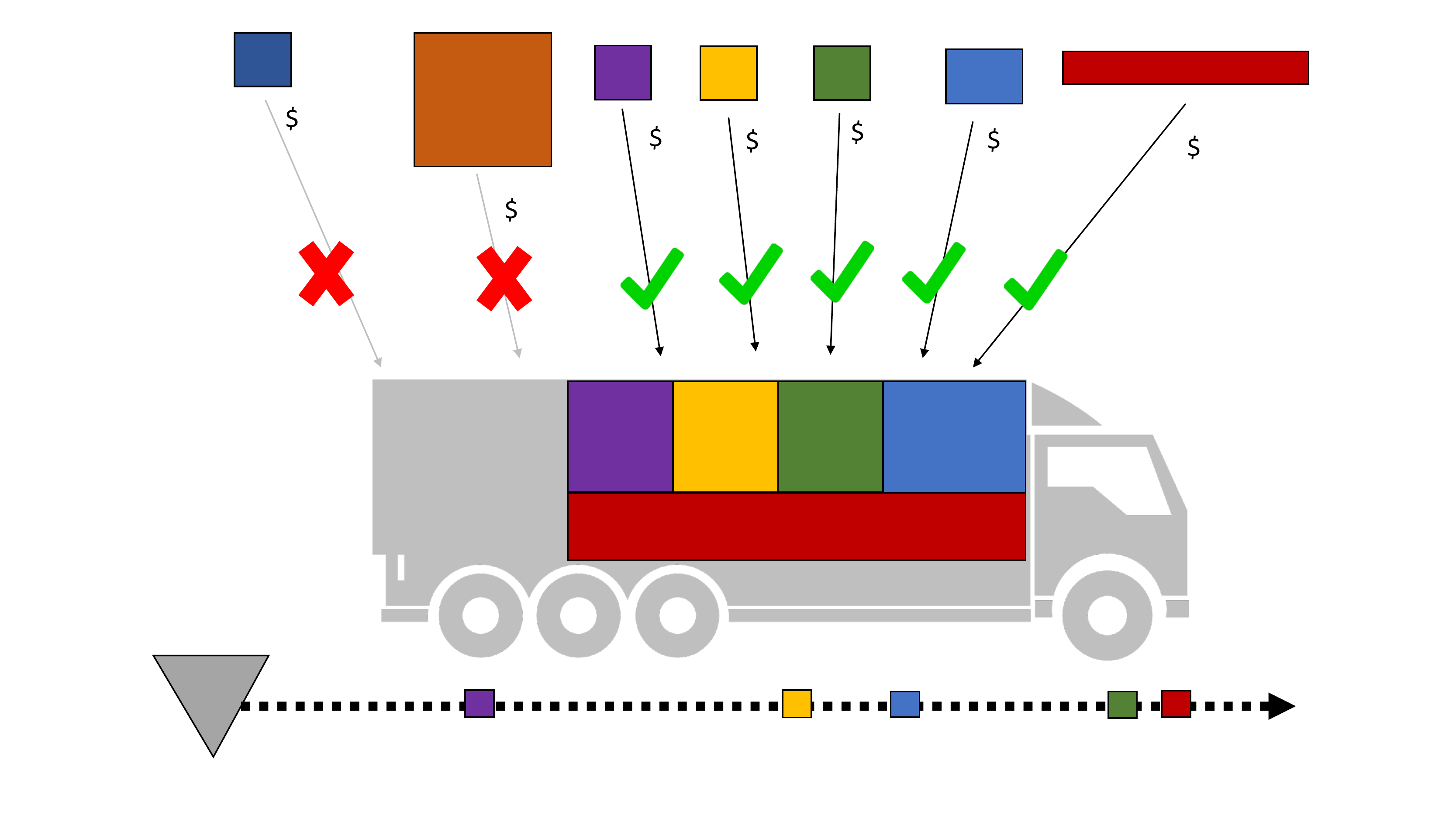}
	\caption{Visual representation of the bidding problem. Modular smart containers (jobs) simultaneously place bids on a transport service with finite capacity; the bids are accepted or rejected based on their marginal contributions.} \label{fig:problem_illustration}
\end{figure}

The cost function and job selection decision of the transport service are defined as well, yet the transport service agent has no learning capacity. As past bids and transport movements do not affect current decisions, the Markovian property is satisfied for this problem. Figure~\ref{fig:problem_illustration} illustrates the bidding problem. 

We now define the jobs, with each job representing a modular container that needs to be transported. A job is represented by the following attribute vector: 

\begin{equation}
\boldsymbol{j} = 
\begin{pmatrix}
j_\tau & = & \text{time till due date} \\
j_d & = & \text{distance to destination} \\
j_v & = & \text{container volume} \notag
\end{pmatrix}
\end{equation}

The attribute $j_\tau \in [0,\tau^{max}] \cup \mathbb{Z}$ indicates how many decision epochs remain until the latest possible shipment date. When a new job enters the system, we set $\mathcal{T}_{\boldsymbol{j}}=\{j_\tau,j_\tau-1,\ldots,0\}$; note that this horizon may differ among jobs and decreases over time; the attribute $j_\tau$ is decremented with each time step. When $j_\tau=0$ and the job has not been shipped, it is considered to be a failed job, incurs a penalty, and is removed from the system. The attribute $j_d \in (0,d^{max}]\cup \mathbb{R}$ indicates the position of the destination on the real line; the further away the higher the transport costs. The job volume $j_v \in [1,\zeta^{max}] \cup \mathbb{Z}$ indicates how much transport capacity the job requires. Let $\boldsymbol{J}_{t^\prime}$ be the problem state, defined as a set containing the jobs present in the system at time ${t^\prime}$. Furthermore, let $\mathcal{J}_{t^\prime}$ be the set of feasible states at time ${t^\prime}$. 

At each decision epoch ${t^\prime}$, a transport service with fixed capacity $C$ departs along the real line. For the transport service to decide which jobs to take, the selection procedure is modeled as a 0-1 knapsack problem that is solved using dynamic programming \cite{kellerer2004}. The value of each job is its bid price minus its transport costs. Jobs with negative values are always rejected. Note that when the transport capacity exceeds the cumulative job volume, the transport service will accept all positive bids. The decision vector for the carrier is denoted as $\boldsymbol{\gamma}=[\gamma_{\boldsymbol{j}}]_{\forall \boldsymbol{j} \in \boldsymbol{J}_{t^\prime}}$, with $\gamma_{\boldsymbol{j}} \in \{0,1\}$. The set $\Gamma(\boldsymbol{J}_{t^\prime})$ denotes the set of all feasible selections. The transport service's cost function for shipping a single job $\boldsymbol{j}$ is a function of distance and volume: $c_{\boldsymbol{j}} = c^{mile} \cdot j_v \cdot j_d$. It maximizes its direct rewards by selecting $\boldsymbol{\gamma}_{t^\prime}$ as follows:
\begin{align}\label{eq:selectioncarrier}
\argmax_{\boldsymbol{\gamma}\in \Gamma(\boldsymbol{J}_{t^\prime})} \left(\sum_{\boldsymbol{j} \in \boldsymbol{J}_{t^\prime}} {\gamma}_{\boldsymbol{j}}(x_{\boldsymbol{j}} - c_{\boldsymbol{j}})\right)\enspace,
\end{align}
s.t.
\begin{align}
\sum_{\boldsymbol{j} \in \boldsymbol{J}_{t^\prime}} {\gamma}_{\boldsymbol{j}} \cdot j_v < C\enspace. \notag
\end{align}

From the perspective of jobs, actions (i.e., bids) are defined on the level of individual containers. All bids $x_{\boldsymbol{j}}\in \mathcal{X}_{\boldsymbol{j}} \equiv \mathbb{R}$ are placed in parallel and independent of one another, yielding a vector $\boldsymbol{x} = [x_{\boldsymbol{j}}]_{\forall \boldsymbol{j} \in \boldsymbol{J}_{t^\prime}}$. Unshipped jobs with $j_\tau>0$ incur holding costs and unshipped jobs with $j_\tau=0$ incur a failed job penalty, both are proportional to the job volume. At any given decision epoch, the direct reward function for individual jobs is defined as follows:
\begin{align}
r_{\boldsymbol{j}}(\gamma_{\boldsymbol{j}},x_{\boldsymbol{j}})= 
\begin{cases}
- x_{\boldsymbol{j}} &  \text{if} \qquad \gamma_{\boldsymbol{j}}=1 \\
- c^{hold} \cdot j_v &  \text{if} \qquad j_\tau>0 \land  \gamma_{\boldsymbol{j}}=0 \\
- c^{pen} \cdot j_v & \text{if} \qquad j_\tau=0 \land  \gamma_{\boldsymbol{j}}=0 \notag
\end{cases}
\end{align}

To obtain $x_{\boldsymbol{j}}$ at the current decision epoch (which may be denoted by $x_{\boldsymbol{j}} \equiv x_{t^\prime,\boldsymbol{j}}$ when explicitly including the decision epoch), we try to solve $\argmax_{x_{t^\prime,\boldsymbol{j}} \in \mathcal{X}_{\boldsymbol{j}}} \\ \mathbb{E}\bigl(\sum_{t^{\prime\prime}=t^\prime}^{t^\prime+|\mathcal{T}_j|} r_{\boldsymbol{j}}(\gamma_{t^{\prime\prime},\boldsymbol{j}},x_{t^{\prime\prime},\boldsymbol{j}})\bigr)$, i.e., the goal is to maximize the expected reward (minimize expected costs) over the horizon of the job. Note that -- as a container cannot observe the bids of other jobs, nor the cost function of the transport service, nor future states -- we can only make decisions based on \textit{expected} rewards depending on the stochastic environment. The solution method presented in Section~\ref{sec:solutionmethod} further addresses this problem.

Finally, we define the transition function for the system state that occurs in the time step from decision epoch $t^\prime$ to $t^\prime+1$. During this step two changes occur; we (i) decrease the due dates of jobs that are not shipped or otherwise removed from the system and (ii) add newly arrived jobs to the state. The set of new jobs arriving for $t^\prime+1$ is defined by $\boldsymbol{\tilde{J}}_{t^\prime+1} \in \mathcal{\tilde{J}}_{t^\prime+1}$. The transition function $S^M:(\boldsymbol{J}_{t^\prime},\boldsymbol{\tilde{J}}_{t^\prime+1},\boldsymbol{\gamma}) \mapsto \boldsymbol{J}_{t^\prime+1}$ is defined by the following sequential procedure:

\setcounter{algo}{0}
\begin{algo}Transition function $S^M(\boldsymbol{J}_{t^\prime},\boldsymbol{\tilde{J}}_{t^\prime+1},\boldsymbol{\gamma}_{t^\prime})$
\end{algo}
\footnotesize
\begin{tabular}{ l  l l}
	\toprule		
	0: & Input: $\boldsymbol{J}_{t^\prime},\boldsymbol{\tilde{J}}_{t^\prime+1},\boldsymbol{\gamma}_{t^\prime}$& $\blacktriangleright$ Current state, job arrivals, shipping selection\\
	1: & $\boldsymbol{J}_{t^\prime+1} \mapsfrom \emptyset$ & $\blacktriangleright$  Initialize next state\\	
	2: & $\boldsymbol{J}_{t^\prime}^x \mapsfrom \boldsymbol{J}_{t^\prime}$& $\blacktriangleright$ Copy state (post-decision state)\\
	3: & $\forall \boldsymbol{j} \in \boldsymbol{J}_{t^\prime}$ & $\blacktriangleright$ Loop over all jobs\\
	4: & \qquad $\boldsymbol{J}_{t^\prime}^x \mapsfrom  \boldsymbol{J}_{t^\prime}^x \setminus \boldsymbol{j}  \mid \gamma_{{t^\prime},\boldsymbol{j}}=1$ & $\blacktriangleright$ Remove shipped job\\
	5: & \qquad $\boldsymbol{J}_{t^\prime}^x \mapsfrom  \boldsymbol{J}_{t^\prime}^x \setminus \boldsymbol{j} \mid j_{\tau}=0 \land \gamma_{{t^\prime},\boldsymbol{j}}=0$ &
	$\blacktriangleright$ Remove unshipped job with due date 0 \\
	6:& \qquad $j_{\tau} \mapsfrom j_{\tau} - 1  \mid j_{\tau}>0$ & $\blacktriangleright$ Decrement time till due date\\
	7: & $\boldsymbol{J}_{{t^\prime}+1} \mapsfrom \boldsymbol{J}_{t^\prime}^x \cup \boldsymbol{\tilde{J}}_{{t^\prime}+1}$ &  $\blacktriangleright$ Merge existing and new job sets \\ 
	8: & Output: $\boldsymbol{J}_{{t^\prime}+1}$ & $\blacktriangleright$ New state\\ 
	\bottomrule  
\end{tabular}
\normalsize
\vspace{4mm}

\section{Solution method}\label{sec:solutionmethod}
To learn the bidding strategy of the containers, we draw from the widely used policy gradient framework. For a detailed description and theoretical background, we refer to the REINFORCE algorithm by Williams \cite{williams1992}. As noted earlier, the policy gradient algorithm returns a stochastic bidding policy, reflecting the deviations in bid prices adopted by individual containers. As bids can take any real value, we must adopt a policy suitable for continuous action spaces. In this paper we opt for a Gaussian policy, drawing bids from a normal distribution. The mean and standard deviation of this distribution are learned using reinforcement learning.

In policy-based reinforcement learning, we perform an action directly on the state and observe the corresponding rewards. Each simulated episode $n \in \{0,1,\ldots,N\}$ yields a batch of selected actions and related rewards according to the stochastic policy $\pi_{\boldsymbol{\theta}}^n(x_{\boldsymbol{j}} \mid {\boldsymbol{j}},\boldsymbol{J}_{t^\prime})=\mathbb{P}^{\boldsymbol{\theta}}(x_{\boldsymbol{j}} \mid \boldsymbol{j},\boldsymbol{J}_{t^\prime})$. Under our Gaussian policy, bids are drawn independently from other containers, i.e., $x_{\boldsymbol{j}} \sim \mathcal{N}(\mu_{\boldsymbol{\theta}}(\boldsymbol{j},\boldsymbol{J}_{t^\prime}), \sigma_{\boldsymbol{\theta}}), \forall \boldsymbol{j} \in \boldsymbol{J}_{t^\prime}$. The randomness in action selection allows the policy to keep exploring the action space and to escape from local optima. From the observed actions and rewards during each episode $n$, we deduce which actions result in above-average rewards and compute gradients ensuring that the policy is updated in that direction, yielding updated policies $\pi_{\boldsymbol{\theta}}^{n+1}$ until we reach $\pi_{\boldsymbol{\theta}}^{N}$. For consistent policy updates, we only use observations for jobs that are either shipped or removed, for which we need some additional notation. Let $\boldsymbol{K}^n=[K_0^n,\ldots,K_{\tau^{max}}^n]$ be a vector containing the number of bid observations for such completed jobs. For example, if a job had an original due date of 4 and is shipped at $j_{\tau}=2$, we would increment $K_{4}^n$, $K_{3}^n$ and $K_{2}^n$ by 1 (using an update function $k(\boldsymbol{j})$). Finally, we store all completed jobs (i.e., either shipped or failed) in a set $\hat{\boldsymbol{J}}^n$. For each episode, the cumulative rewards per job are defined  as follows:
\begin{align}
\hat{v}_{t,\boldsymbol{j}}^n(\gamma_{\boldsymbol{j}},x_{\boldsymbol{j}})= 
\begin{cases}
r_{\boldsymbol{j}}(\gamma_{\boldsymbol{j}},x_{\boldsymbol{j}}) &  \text{if} \qquad t=0 \\
r_{\boldsymbol{j}}(\gamma_{\boldsymbol{j}},x_{\boldsymbol{j}})+\hat{v}_{t-1,\boldsymbol{j}}^n & \text{if} \qquad t>0 \notag
\end{cases}
\quad, \forall t \in \mathcal{T}_{\boldsymbol{j}}\enspace.
\end{align}
Let $\hat{\boldsymbol{v}}_{t^\prime}^n=\bigl[[\hat{v}_{t,\boldsymbol{j}}^n]_{t \in \mathcal{T}_j}\bigr]_{\forall \boldsymbol{j} \in \boldsymbol{J}_t}$ be the vector containing all observed cumulative rewards at time $t^\prime$ in episode $n$. At the end of each episode, we can then define the information vector

\begin{align}
\boldsymbol{I}^n =\biggl[[\boldsymbol{J}_{t^\prime}^n, \boldsymbol{x}_{t^\prime}^n, \hat{\boldsymbol{v}}_{t^\prime}^n,
\boldsymbol{\gamma}_{t^\prime}^n]_{\forall t^\prime \in \{0,\ldots,T\}}, 
\boldsymbol{K}^n, \hat{\boldsymbol{J}}^n\biggr] \notag
\end{align}

\noindent and corresponding updating function $i(\cdot)$; the information vector contains the states, actions and rewards necessary for the policy updates (i.e., a history similar to the well-known SARSA trajectory). The decision-making policy is updated according to the policy gradient theorem \cite{sutton2018}, which may be formalized as follows:
\begin{align}	
\nabla_{\boldsymbol{\theta}} v_{j_\tau,\boldsymbol{j}}^{\pi_{\boldsymbol{\theta}}}
\propto \sum_{{t^\prime}=1}^{T} \left(\int_{\boldsymbol{J}_{t^\prime} \in \mathcal{J}_{t^\prime}} \mathbb{P}^{\pi_{\boldsymbol{\theta}}}(\boldsymbol{J}_{t^\prime} \mid \boldsymbol{J}_{{t^\prime}-1}) \int_{x_{\boldsymbol{j}} \in \mathcal{X}_{\boldsymbol{j}}}   \nabla_{\boldsymbol{\theta}}{\pi_{\boldsymbol{\theta}}}({\boldsymbol{x}_{\boldsymbol{j}}} \mid \boldsymbol{j}, \boldsymbol{J}_{t^\prime})v_{j_\tau,\boldsymbol{j}}^{\pi_{\boldsymbol{\theta}}}\right)  \enspace.  \notag
\end{align}
Essentially, the theorem states that the gradient of the objective function is proportional to the value functions $v_{j_\tau,\boldsymbol{j}}^{\pi_{\boldsymbol{\theta}}}$ multiplied by the policy gradients for all actions in each state, given the probability measure $\mathbb{P}^{\pi_{\boldsymbol{\theta}}}$ implied by the prevailing decision-making policy $\pi_{\boldsymbol{\theta}}$.

We proceed to formalize the policy gradient theorem for our problem setting. Let $\boldsymbol{\theta}$ be a  vector of weight parameters that defines the decision-making policy $\pi_{\boldsymbol{\theta}}: (\boldsymbol{j},\boldsymbol{J}_{t^\prime}) \mapsto \mathcal{X}_{\boldsymbol{j}}$. Furthermore, let $\boldsymbol{\phi}(\boldsymbol{j},\boldsymbol{J}_{t^\prime})$ be a feature vector that distills the most salient attributes from the problem state, e.g., the number of jobs waiting to be transported or the average time till due date. We will further discuss the features in Section~\ref{sec:numericalexperiments}. For the Gaussian case, we formalize the policy as follows:
\begin{align}
\pi_{\boldsymbol{\theta}}=\frac{1}{\sqrt{2\pi}\sigma_{\boldsymbol{\theta}}}e^{-\frac{\left(x_{\boldsymbol{j}}-\mu_{\boldsymbol{\theta}}\left(\boldsymbol{j},\boldsymbol{J}_{t^\prime}\right)\right)^2}{2\sigma_{\boldsymbol{\theta}}^2}}\enspace, \notag
\end{align}
with $x_{\boldsymbol{j}}$ being the bid price, $\mu_{\boldsymbol{\theta}}(\boldsymbol{j},\boldsymbol{J}_{t^\prime}) = \boldsymbol{\phi}(\boldsymbol{j},\boldsymbol{J}_{t^\prime})^\top {\boldsymbol{\theta}}$ the Gaussian mean and $\sigma_{\boldsymbol{\theta}}$ the parametrized standard deviation. The action $x_{\boldsymbol{j}}$ may be obtained from the inverse normal distribution. The corresponding gradients are defined by
\begin{align}
\nabla_{\mu_{\boldsymbol{\theta}}}(\boldsymbol{j},\boldsymbol{I}^n) = \nabla_{\boldsymbol{\theta}}(\boldsymbol{j},\boldsymbol{I}^n)  = \\ \nabla_{\boldsymbol{\theta}} \log \pi_{\boldsymbol{\theta}} (
\boldsymbol{j},\boldsymbol{I}^n) =\\  \frac{(x_{\boldsymbol{j}}-\mu_{\boldsymbol{\theta}}(\boldsymbol{j},\boldsymbol{J}_{t^\prime}))\phi(\boldsymbol{j},\boldsymbol{J}_{t^\prime})}{\sigma_{\boldsymbol{\theta}}^2} \enspace, \notag
\end{align}

\begin{align}
\nabla_{\sigma_{\boldsymbol{\theta}}} (\boldsymbol{j},\boldsymbol{I}^n)  = \frac{(x_{\boldsymbol{j}}-\mu_{\boldsymbol{\theta}}(\boldsymbol{j},\boldsymbol{J}_{t^\prime}))^2 - \sigma_{\boldsymbol{\theta}}^2}{\sigma_{\boldsymbol{\theta}}^3}\enspace. \notag
\end{align}

The gradients are used to update the policy parameters. As the observations may exhibit large variance, we add a non-biased baseline value (i.e., not directly depending on the policy), namely the average observed value during the episode \cite{sutton2018}. For the prevailing episode $n$, we define the baseline as 
\begin{align}
\bar{v}_t^n = \frac{1}{K_{t}^n} \sum_{\boldsymbol{j} \in \boldsymbol{\hat{J}}^n} \hat{v}_{t,\boldsymbol{j}}^n, \forall t \in \{0,\ldots, \tau^{max}\}  \enspace. \notag
\end{align}
For the Gaussian policy, the weight update rule for $\mu_{\boldsymbol{\theta}}$ (updating $\boldsymbol{\theta}^{n-1}$ to $\boldsymbol{\theta}^n$) is:
\begin{align}\label{eq:update_mu}
\Delta_{\mu_{\boldsymbol{\theta}}}(\boldsymbol{j},\boldsymbol{I}^n)   = \Delta_{\boldsymbol{\theta}}(\boldsymbol{j},\boldsymbol{I}^n)  = \\
 \alpha_\mu \frac{1}{K_{j_\tau}^n}(\hat{v}_{j_\tau} - \bar{v}_{j_\tau}) \nabla_{\boldsymbol{\theta}} \log \pi_{\boldsymbol{\theta}} (\boldsymbol{j},\boldsymbol{I}^n)  =\\  
\alpha_\mu \frac{1}{K_{j_\tau}^n}(\hat{v}_{j_\tau} - \bar{v}_{j_\tau}) \left[\frac{(x_{\boldsymbol{j}}-\mu_{\boldsymbol{\theta}}(\boldsymbol{j},\boldsymbol{J}_{t^\prime}))\phi(\boldsymbol{j},\boldsymbol{J}_{t^\prime})}{\sigma_{\boldsymbol{\theta}}^2}\right]\enspace. 
\end{align}
The standard deviation $\sigma_{\boldsymbol{\theta}}$ is updated as follows:
%
\begin{align}\label{eq:update_stdev}
\Delta_{\sigma_{\boldsymbol{\theta}}}(\boldsymbol{j},\boldsymbol{I}^n) = \alpha_\sigma \frac{1}{K_{j_\tau}^n} (\hat{v}_{j_\tau} - \bar{v}_{j_\tau}) \nabla_{\boldsymbol{\theta}} \log \pi_{\boldsymbol{\theta}} (
\boldsymbol{j},\boldsymbol{I}^n)= \\ 
\alpha_\sigma  \frac{1}{K_{j_\tau}^n}(\hat{v}_{j_\tau} - \bar{v}_{j_\tau}) \left[\frac{(x_{\boldsymbol{j}}-\mu_{\boldsymbol{\theta}}(\boldsymbol{j},\boldsymbol{J}_{t^\prime}))^2 - \sigma_{\boldsymbol{\theta}}^2}{\sigma_{\boldsymbol{\theta}}^3}
\right]\enspace. 
\end{align}
Intuitively, this means that after each episode we update the feature weights -- which in turn provide the state-dependent mean bidding price -- and the standard deviation of the bids. The mean bidding price -- taking into account both individual job properties and the state of the system -- represents the bid that is expected to minimize overall costs. If effective bids are very close to the mean, the standard deviation will decrease and the bidding policy will converge to an almost deterministic policy. However, if there is an expected benefit in varying bids, the standard deviation may grow larger. The algorithmic outline to update the parametrized policy is at follows:

\setcounter{algo}{1}
\begin{algo}\label{policygradientoutline}Outline of the policy gradient bidding algorithm (based on \cite{williams1992})
\end{algo}
\begin{tabular}{ l l l}
	\toprule		
	0: & Input:  $\pi_{\boldsymbol{\theta}}^0$ & $\blacktriangleright$ Differentiable parametrized policy\\	
	1: & $\alpha_\mu \mapsfrom (0,1), \alpha_\sigma \mapsfrom (0,1)$ & $\blacktriangleright$ Set step sizes\\
	2: & $\sigma^0 \mapsfrom \mathbb{R}^+$ & $\blacktriangleright$ Initialize standard deviation\\
	3: & $\boldsymbol{\theta} \mapsfrom \mathbb{R}^{|\boldsymbol{\theta}|}$ & $\blacktriangleright$ Initialize policy parameters\\
	4: & $\forall \boldsymbol{n} \in \{0,\ldots,N\}$ & $\blacktriangleright$ Loop over episodes\\
	5: & \qquad $\boldsymbol{\hat{J}}^n \mapsfrom \emptyset$ & $\blacktriangleright$ Initialize completed job set\\
	6: & \qquad $\boldsymbol{I}^n \mapsfrom \emptyset$ & $\blacktriangleright$ Initialize information set\\
	7: & \qquad $\boldsymbol{J}_0 \mapsfrom \mathcal{J}_0$ & $\blacktriangleright$ Generate initial state\\
	8: & \qquad $\forall t^\prime \in \{0,\ldots,T\}$ & $\blacktriangleright$ Loop over finite time horizon\\
	9: & \qquad\qquad  $x_{\boldsymbol{j}}^n \mapsfrom \pi_{\boldsymbol{\theta}}^n(\boldsymbol{j},\boldsymbol{J}_{t^\prime}), \forall \boldsymbol{j} \in \boldsymbol{J}_{t^\prime}$ & $\blacktriangleright$ Bid placement jobs\\
	10: & \qquad\qquad  $\boldsymbol{\gamma}^n \mapsfrom \argmax_{\boldsymbol{\gamma}^n \in \Gamma(\boldsymbol{J}_{t^\prime})} $ & $\blacktriangleright$ Job selection carrier, Eq. \eqref{eq:selectioncarrier}\\
	& \qquad\qquad $\left(\sum_{\boldsymbol{j} \in \boldsymbol{J}_{t^\prime}} {\gamma}_{\boldsymbol{j}}^n(x_{\boldsymbol{j}} - c_{\boldsymbol{j}})\right)$ & \\
	11: & \qquad\qquad $\hat{v}_{j_\tau,\boldsymbol{j}}^n \mapsfrom r_{\boldsymbol{j}}(\gamma_{\boldsymbol{j}}^n,x_{\boldsymbol{j}}^n), \forall \boldsymbol{j} \in \boldsymbol{J}_{t^\prime}$ & $\blacktriangleright$ Compute cumulative rewards\\ 
	12: & \qquad\qquad $\forall \boldsymbol{j} \in \boldsymbol{J}_{t^\prime} \mid j_{\tau}=0 \lor {\gamma}_{\boldsymbol{j}}^n=1 $& $\blacktriangleright$ Loop over completed jobs \\
	13: & \qquad\qquad\qquad $\boldsymbol{\hat{J}}^n \mapsfrom \boldsymbol{\hat{J}}^n \cup \{\boldsymbol{j}\}$ & $\blacktriangleright$ Update set of completed jobs\\ 
	14: & \qquad\qquad\qquad $\boldsymbol{K}^n \mapsfrom k(\boldsymbol{j})$ & $\blacktriangleright$ Update number of completed jobs\\ 
	15: & \qquad\qquad $\boldsymbol{I}^n \mapsfrom i\biggl(\boldsymbol{J}_{t^\prime}^n, \boldsymbol{x}_{t^\prime}^n, \hat{\boldsymbol{v}}_{t^\prime}^n,
	\boldsymbol{\gamma}_{t^\prime}^n, \boldsymbol{K}^n, \hat{\boldsymbol{J}}^n, \boldsymbol{I}^n\biggr)$ & $\blacktriangleright$ Store information\\
	16: & \qquad\qquad $\boldsymbol{\tilde{J}}_{t^\prime}\mapsfrom \mathcal{\tilde{J}}_{t^\prime}$ & $\blacktriangleright$ Generate job arrivals\\
	17: & \qquad\qquad $\boldsymbol{J}_{t^\prime+1} \mapsfrom S^M(\boldsymbol{J}_{t^\prime},\boldsymbol{\tilde{J}}_{t^\prime+1},\boldsymbol{\gamma}_{t^\prime}^n) $ & $\blacktriangleright$ Transition function, Algorithm 1\\
	18: & \qquad $\forall t \in \{0,\ldots,\tau^{max}\}$ & $\blacktriangleright$ Loop till maximum due date\\
	19: & \qquad\qquad $\forall \boldsymbol{j} \in \boldsymbol{\hat{J}}^n$ & $\blacktriangleright$ Loop over completed jobs\\
	20: & \qquad\qquad\qquad $\mu_{\boldsymbol{\theta}}^{n+1} \mapsfrom \mu_{\boldsymbol{\theta}}^n+\Delta_{\mu_{\boldsymbol{\theta}}}(\boldsymbol{j},\boldsymbol{I}^n) $ & $\blacktriangleright$ Update Gaussian mean, Eq. \eqref{eq:update_mu} \\
	21:& \qquad\qquad\qquad $\sigma_{\boldsymbol{\theta}}^{n+1} \mapsfrom \sigma_{\boldsymbol{\theta}}^n+\Delta_{\sigma_{\boldsymbol{\theta}}}(\boldsymbol{j},\boldsymbol{I}^n) $ & $\blacktriangleright$ Update standard deviation, Eq. \eqref{eq:update_stdev}\\
	22: & Output: $\pi_{\boldsymbol{\theta}}^N$ & $\blacktriangleright$ Return tuned policy\\
	\bottomrule  
\end{tabular}

\section{Numerical experiments}\label{sec:numericalexperiments}
This section describes the numerical experiments and the results. Section~\ref{ssec:exploration} explores the parameter space and aids in tuning the hyperparameters. The algorithm is written in Python 3.7 and available online.\footnote{https://github.com/woutervanheeswijk/policygradientsmartcontainers}

\subsection{Exploration of parameter space}\label{ssec:exploration}
The purpose of this section is twofold: we explore the effects of various parameter settings on the performance of the algorithm and select a suitable set of parameters for the remainder of the experiments. We make use of the instance settings summarized in Table~\ref{table:instancesettings}. Note that the penalty for failed jobs is the main driver in determining bid prices; together with holding costs, it intuitively represents the maximum price the smart container is willing to bid to be transported. 
\begin{table}
	\centering
	\caption{Instance settings}
	\label{table:instancesettings}
\begin{tabular}{ l  l }
	\toprule		
	  Max. \# job arrivals & [0-10] \\
	  Due date & [1-5] \\
	  Job transport distance & [10-100] \\
	  Job volume & [1-10]\\
	  Holding cost (per volume unit) & 1\\
	  Penalty failed job (per volume unit) & 10\\
	  Transport costs per mile (per volume unit) & 0.1 \\
	  Transport capacity & 80 \\
	\bottomrule  
\end{tabular}
\end{table}

To parametrize the policy we use several features. First, we use a scalar that serves as the bias. Second, we use the individual job properties of the job placing the bid, i.e., the time till due date, the job's transport distance and the container volume. Third, in case the job shares its own properties with the system, it also use the generic system features: the total number of jobs, average distance, total volume, and average due date. Recall that these system features only include the data of other smart containers that share their information. All weight parameters in $\boldsymbol{\theta}$ are initialized at 0, yielding initial bid prices of 0.

\begin{figure}
	\centering
	\begin{subfigure}{.5\textwidth}
		\centering
		\includegraphics[width=1.0\linewidth]{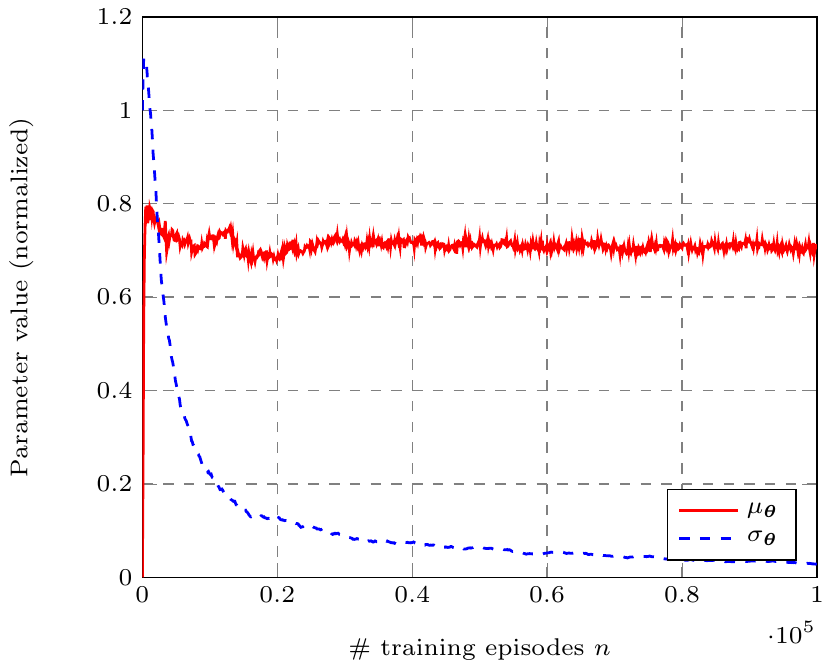}
		\caption{$\alpha_\mu=0.01$ and $\alpha_\sigma=0.001$}
		\label{fig:small_learning_rates}
	\end{subfigure}%
	\begin{subfigure}{.5\textwidth}
		\centering
		\includegraphics[width=1.0\linewidth]{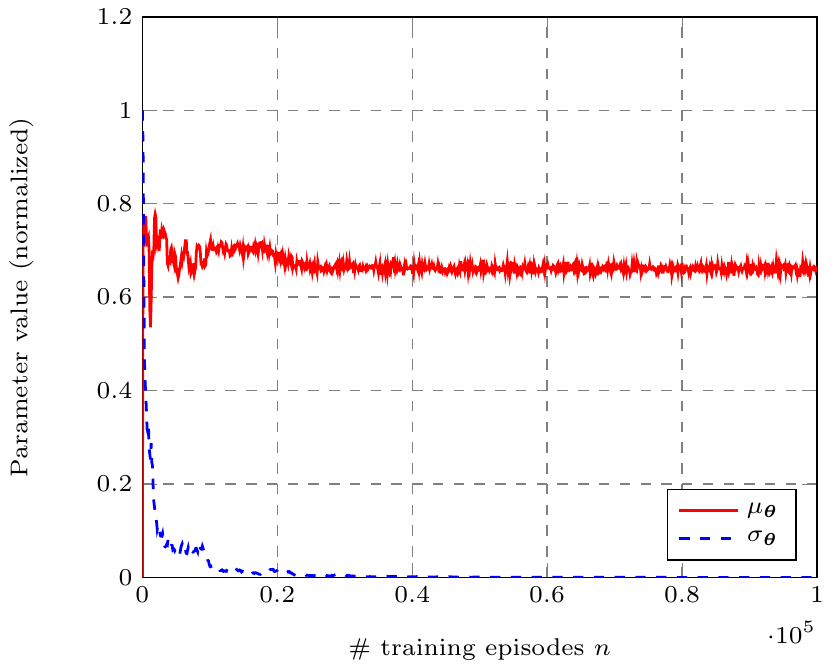}
		\caption{$\alpha_\mu=0.1$ and $\alpha_\sigma=0.01$}
		\label{fig:large_learning_rates}
	\end{subfigure}
	\caption{Convergence of $\mu_{\boldsymbol{\theta}}$ and $\sigma_{\boldsymbol{\theta}}$ (normalized) for various learning rates. Higher learning rates achieve both faster convergence and lower average bid prices.}
	\label{fig:convergence_learning_rates}
\end{figure}

We perform a sequential search to set the simulation parameters. First, we tune the learning rates $\alpha_\mu$ (learning rate for mean) and $\alpha_\sigma$ (learning rate for standard deviation), starting with a standard normal distribution. We test learning rates $\{0.0001, 0.001, 0.01, 0.1\}$ for both parameters and find that $\alpha_\mu=0.1$ and $\alpha_\mu=0.01$ are stable (i.e., no exploding gradients) and converge reasonably fast. Taking smaller learning rates yields no eminent advantages in terms of stability or eventual policy quality. 
Figure~\ref{fig:convergence_learning_rates} shows two examples of parameter convergence under various learning rates.

Next, we tune the initial bias weight $\theta_0^0$ (using values $\{-50,-40,\ldots,40,50\}$) and the initial standard deviation $\sigma^0$ (using values $\{0.01, 0.1, 1, 10,25\}$). Anticipating non-zero bids, we test several initializations with nonzero bias weights. Large standard deviations allow for much exploration early on, but may also result in unstable policies or slow convergence. From the experiments, we observe that the bias weight converges to a small or negative weight and that there is no benefit in different initializations. For the standard deviation, we find that $\sigma^0 = 10$ yields the best results; the exploration triggered by setting large initial standard deviations helps avoiding local optima early on. In terms of performance, the average transport costs are 7.3\% lower than under the standard normal initialization. Standard deviations ultimately converge to similarly small values regardless the initialization.

\begin{figure}
	\centering
	\begin{subfigure}{.5\textwidth}
		\centering
		\includegraphics[width=1.0\linewidth]{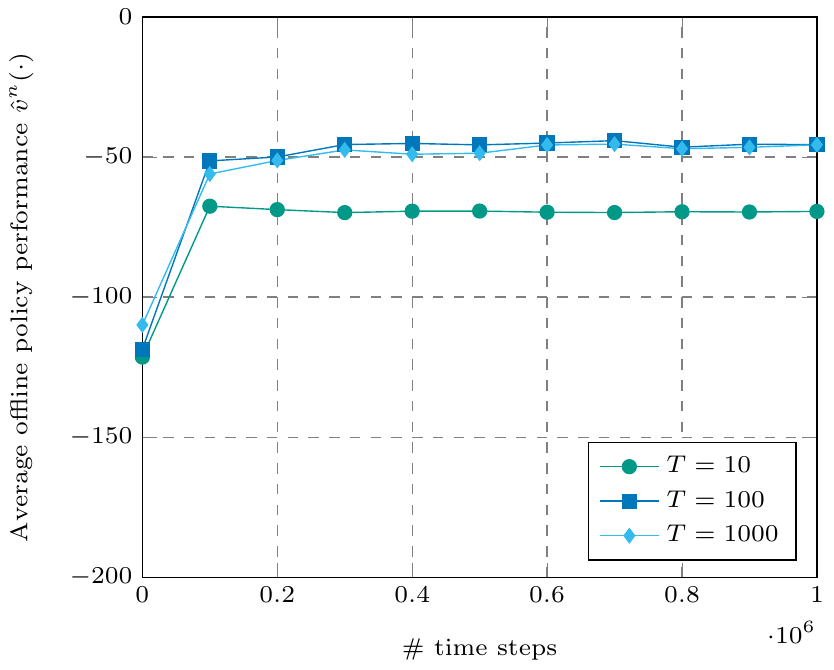}
		\caption{Comparison offline quality.} 
		\label{fig:offlineperformance}
	\end{subfigure}%
	\begin{subfigure}{.5\textwidth}
		\centering
		\includegraphics[width=1.0\linewidth]{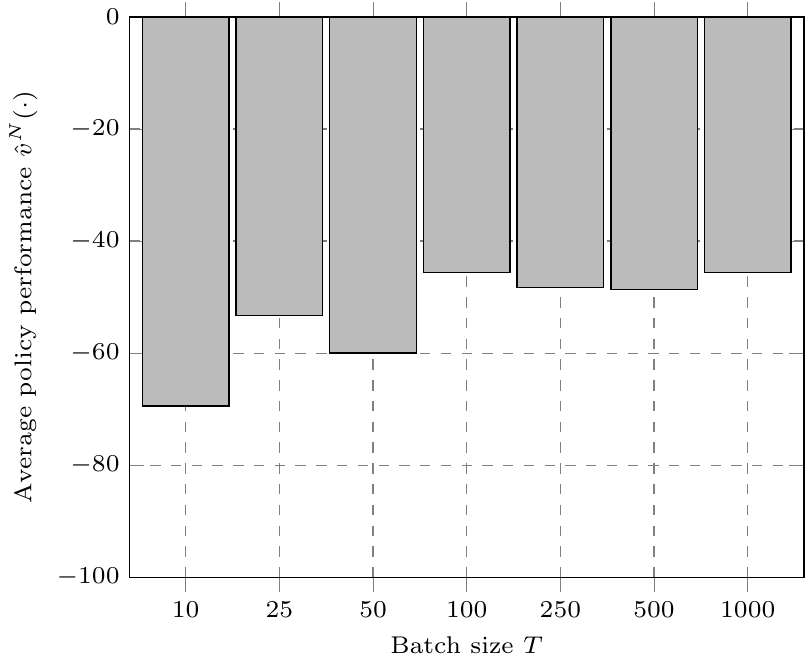}
		\caption{Comparison final policy quality.}
		\label{fig:largebatch}
	\end{subfigure}
	\caption{Policy performance for various time horizons. The horizon $T=100$ yields the best overall performance; batches too small diminish performance.}
	\label{fig:figure_episodes}
\end{figure}

Finally, we determine the number of episodes $N$ and the length of each horizon $T$. Longer horizons lead to larger and therefore more reliable batches of completed jobs per episode, but naturally require more computational effort. Thus, we compare settings for which the total number of time steps $N \cdot T$ is equivalent. Each alternative simulates 1,000,000 time steps, using $T=\{10,25,50,100,250,500,1000\}$ with corresponding values $N$. To test convergence, after each 10\% of completed training episodes we perform 10 validation episodes -- always with $T=1000$ for fair comparisons -- to evaluate policy qualities. We find that having large batches provides notable advantages. Furthermore, in all cases 400,000 time steps appear sufficient to converge to stable policies. To illustrate the findings, Figure~\ref{fig:offlineperformance} shows the average transport costs measured after each 100,000 time steps (using the then-prevailing policy); Figure~\ref{fig:largebatch} shows the quality of the eventual policies for each time horizon. 

The final parameters to be used for the remainder of the numerical experiments are summarized as follows: $N=4,000$, $T=100$, $\sigma^0=10$, $\alpha_\mu=0.1$ and $\alpha_\sigma=0.01$.

\subsection{Analysis of policy performance}
Having determined the parameter settings, we proceed to analyze the performance of the jointly learned policies. This section addresses the effects of information sharing, the relevance of the used features in determining the bid, and the behavior of the bidding policy and its impact on carrier profits. All results in this section correspond to the performance of policies after training is completed. To obtain additional insights we sometimes define alternative instances; for conciseness we do not explicitly define the settings of these alternatives.

We first evaluate the effects of information sharing. According to preset ratios, we randomly assign whether or not a container shares its information with the system. Only containers that share information are able to see aggregate system information. Clearly, the more containers participate, the more accurate the system information will be. We test sharing rates ranging from 0\% to 100\% with increments of 10\%; Table~\ref{table:featureweights} shows the results. We observe that performance under full information sharing and no information sharing is almost equivalent, with partial information sharing always performing worse. The latter observation may be explained by the distorted view presented by only measuring part of the system state.

\begin{table}
	\scriptsize
	\centering
	\caption{Feature weights for various information sharing rates.}
	\label{table:featureweights}
	\begin{tabular}{ l | c| c| c| c| c| c |c| c| c |c| c }
		\hline
		Feature &0\% &10\%&20\%&30\%&40\%&50\%&60\%&70\%&80\%&90\%&100\%\\
		\hline		
		Scalar & -10.10  & -10.28&-8.03&-11.83&-10.98&-4.81&-7.17&-10.46&-8.71&-9.46&-12.12\\
		Total job volume & --  &0.06&0.06&-0.12&0.44&0.03&0.29&0.42&1.18&0.64&0.63\\
		Average due date & -- &-2.37&-2.73&-3.28&-2.57&-2.73&-2.69&-2.49&-1.79&-2.02&-1.56\\
		Average distance & -- &1.83&3.07&2.59&2.36&3.18&3.58&3.16&3.44&2.26&2.92\\
		\# jobs & --  &-0.03&-0.31&-0.69&-0.21&-0.86&-0.77&-1.39&-0.02&-1.25&-1.70\\
		Job volume   & 59.20  & 57.14&56.83&58.46&58.70&55.27&56.13&56.55&56.69&57.97&60.10\\
		Job due date & -22.45  & -23.25&-24.02&-21.16&-21.95&-23.89&-25.96&-22.89&-22.26&-22.82&-21.27\\
		Job distance & 49.49 &48.37&45.88&49.38&48.81&45.98&45.27&46.85&48.00&49.15&49.54\\
		\hline
		\textbf{Average reward} & \textbf{-46.87}  &\textbf{-51.78}&\textbf{-55.96}&\textbf{-49.12}&\textbf{-49.54}&\textbf{-57.75}&\textbf{-56.75}&\textbf{-55.71}&\textbf{-53.12}&\textbf{-49.43}&\textbf{-46.32}\\
		\hline  
	\end{tabular}
\end{table}

\begin{figure}
	\centering
	\begin{subfigure}{.5\textwidth}
		\centering
		\includegraphics[width=1.0\linewidth]{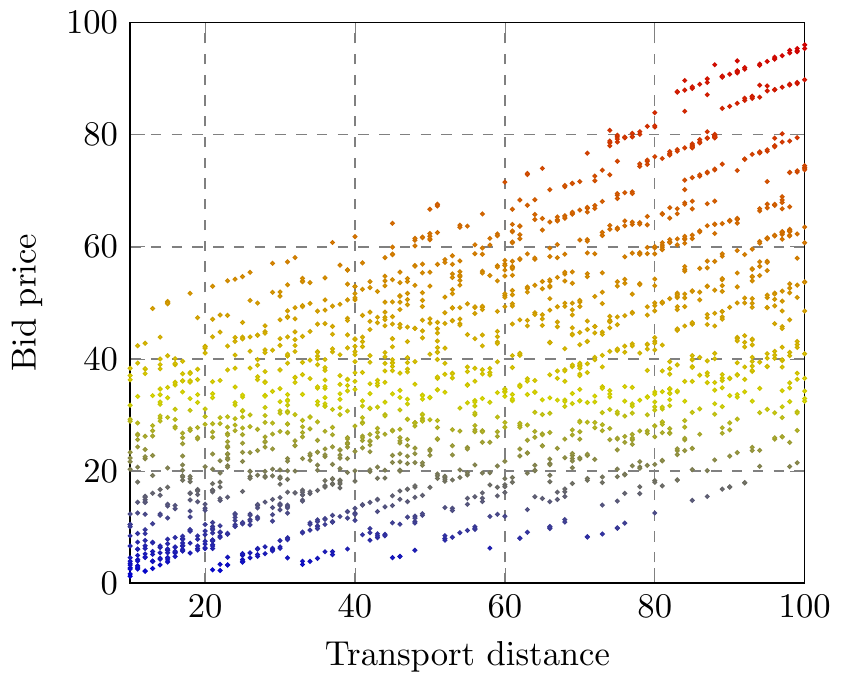}
		\caption{Bids relative to transport distance.} 
		\label{fig:bid_cloud}
	\end{subfigure}%
	\begin{subfigure}{.5\textwidth}
		\centering
		\includegraphics[width=1.0\linewidth]{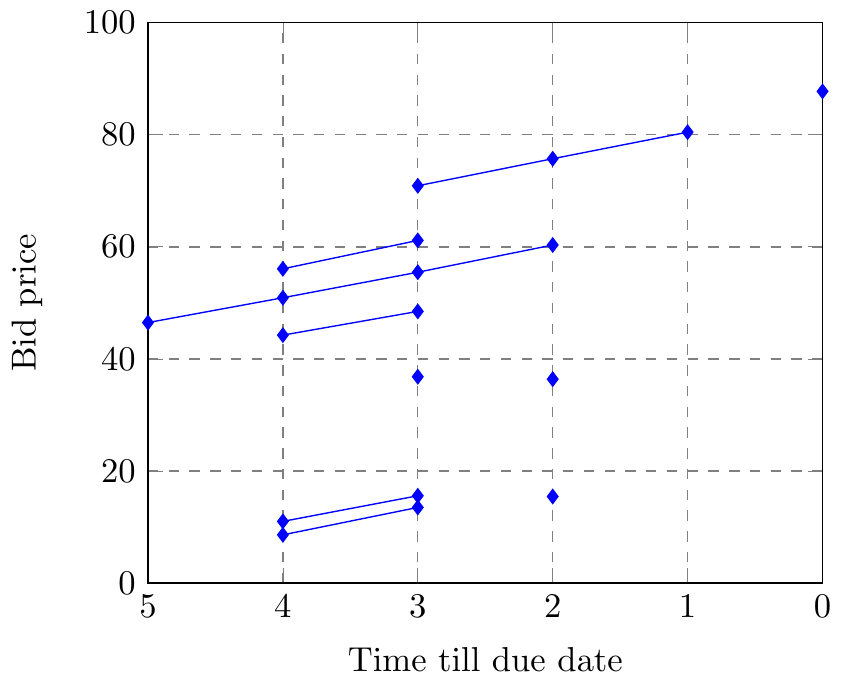}
		\caption{Sample paths of bids over time.}
		\label{fig:bids_over_time}
	\end{subfigure}
	\caption{Visualizations of bidding policies with respect to volume and due date Bids tend to increase when (a) transport distance is larger and (b) the job is closer to its due date.}
	\label{fig:bidding_behavior}
\end{figure}

In the policy parametrization all features are scaled to the $[0,1]$ domain, such that the magnitudes of weights are comparable among each other. We see that the generic features have relatively little impact on the overall bid price. This underlines the limited difference observed between full information sharing and no information sharing. Job volume and job distance are by far the most significant drivers of a job's bid prices. In line with expectations, as the costs incurred by the carrier depend on these two factors, therefore requiring higher compensations. Furthermore, holding- and penalty costs are proportional to the job volume. The relationship with the time till due date is negative; if more time remains, it might be prudent to place lower bids without an imminent risk of penalties. On average, each job places 1.36 bids and 99.14\% of jobs is ultimately shipped; the capacity of the transport service is rarely restrictive. Figure~\ref{fig:bidding_behavior} illustrates bidding behavior with respect to transport distance and time till due date, respectively.

Next, we discuss some behavior of the bidding policy and its effect on carrier profits. As our carrier is a passive agent we omit overly quantitative assessments here. We simulate various toy-sized instances, adopting simplified deterministic settings with a single container type (time till due date is zero, identical volume and distance). If transport capacity is guaranteed to suffice, the learned bid prices converge to almost zero. If two jobs always compete for one transport service and the other incurs a penalty, the bid will be slightly below the penalty cost. Several other experiments with scarce capacity also show bid prices slightly below the expected penalty costs that would otherwise be incurred. For our base scenario, the profit margin for the carrier is 20.2\%. This positive margin indicates that the features do not encapsulate all information required to learn the minimum bidding price. For comparison, we run another scenario in which the carrier's transport costs -- which are unknown to the smart containers -- are the sole feature; in that case all jobs trivially learn the minimum bidding price. This result implies that carrier may have financial incentives not to divulge too much information to the smart containers.

For the carrier, the bidding policies deployed by the smart container greatly influence its profit. Scarcity of transport capacity drives up bid prices, yet also increases the likelihood of missed revenue. To gain more insight into this trade-off, we simulate various levels of transport capacity, from scarce to abundant. These experiment indeed confirm that a (non-trivial) capacity level exists that maximizes profit. In addition, a carrier needs not to accept all jobs whose bid exceed their marginal transport costs, as we presumed in this paper. Having a carrier represented by an active agent stretches beyond the scope of this paper. 

We summarize the main findings, reiterating that the setting of this paper is a highly stylized environment. The key insights are as follows:

\begin{itemize}
\item Utilizing global system information only marginally reduces job's transport costs compared to sharing no system information;
\item Jointly learned policies converge to stable bidding policies with small standard deviations;
\item Jobs with more time remaining till their due date are prone to place lower bids;
\item Carriers have an incentive not to disclose true cost information when transport capacity is abundant.
\end{itemize}

\section{Conclusions}\label{sec:conclusions}
Traditional transport markets rely on (long-term) contracts between shippers and carriers in which price agreements are made. In contrast, self-organizing systems are expected to evolve into some form of spot market where demand and supply are dynamically matched based on the current state of the system. This paper explores the concept of smart containers placing bids on restricted transport capacity. We design a policy gradient algorithm to mimic joint learning of a bidding policy, sharing observations between autonomous smart modular containers. The stochastic policy reflects deviations made by individual containers, given that deterministic policies are easy to counter in a competitive environment. This stochastic approach appears effective. Standard deviations converge to small values, implying stable bidding policies. The performance of the policy is consistent with the effects of job volume, transport distance and till time due date, which are used to parametrize the bidding policy.

Numerical experiments show that sharing system information only marginally reduces bidding costs; individual job properties are the main driver in setting bids. The limited difference in policy quality with and without sharing system information is an interesting observation. This result implies that smart containers would only need to (anonymously) submit their key properties, submitted bid prices and incurred costs. There is no apparent need to share information on a system-wide level, which would greatly ease the system design.

The profitability of the transport service -- which is modeled as a passive agent in this paper -- strongly depends on the bidding policy of the smart containers. Experiments with varying transport capacities show that in turn, the carrier can also influence bidding policies by optimizing the transport capacity that is offered. Without central governance, unbalances between smart containers and transport services may cause disturbed performances. Based on the findings presented in this paper, one can imagine that the dynamic interplay between carriers and smart containers is a very interesting one that begs closer attention.

We re-iterate that this study is of an explorative nature; there are many avenues for follow-up research. In terms of algorithmic improvements, actor-critic methods (learning functions for expected downstream values rather than merely observing them) would be a logical continuation. Furthermore, the linear expression to compute the bidding price could be replaced by neural networks that capture potential non-linear structures. In addition, the basic problem presented in this paper lends itself for various extensions. So far the carrier has been assumed to be a passive agent, offering fixed transport capacity and services regardless of (anticipated) income. In reality the carrier will also make intelligent decisions based on the bidding policies of smart containers. A brokerage structure might also emerge in the Physical Internet context. Finally, we considered only a single transport service operating on the real line. Using the same algorithmic setup, this setting could be extended to more realistic networks with multiple carriers, routes and destination nodes. 

 \bibliographystyle{apacite}
 \bibliography{bibliographyphysicalinternet}

\end{document}